%% file: main.tex
\documentclass[a4paper, 10pt, conference]{ieeeconf}      

\IEEEoverridecommandlockouts                              
\usepackage[pdftex]{graphicx} 

\usepackage{amsmath} 
\usepackage{amssymb}  
\usepackage{bm}  
\usepackage{color}
\usepackage{amsthm}
\usepackage{siunitx}
\usepackage{hyperref}
\usepackage{cite}
\usepackage{balance}

\begin{document}

\title{\LARGE \bf \textbf{\texttt{RPC}}: A Modular Framework for Robot Planning, Control, and Deployment}

\author{Seung Hyeon Bang$^{1}$, Carlos Gonzalez$^{1}$, Gabriel Moore$^{2}$, Dong Ho Kang$^{3}$, Mingyo Seo$^{2}$ and Luis Sentis$^{1}$
\thanks{$^{1}$S.H. Bang, C. Gonzalez, and L. Sentis are with the Department of Aerospace Engineering and Engineering Mechanics, The University of Texas at Austin, TX 78712, USA
         \tt\small bangsh0718@utexas.edu}%
 \thanks{$^{2}$G. Moore and M. Seo are with the Department of Electrical and Computer Engineering, The University of Texas at Austin, TX 78712, USA}%
 \thanks{$^{3}$D.H. Kang is with the Department of Mechanical Engineering, The University of Texas at Austin, TX 78712, USA}}%

\maketitle

\thispagestyle{empty}
\pagestyle{empty}



\begin{abstract}
This paper presents an open-source, lightweight, yet comprehensive software framework, named \textbf{\texttt{RPC}}, which integrates physics-based simulators, planning and control libraries, debugging tools, and a user-friendly operator interface. \textbf{\texttt{RPC}} enables users to thoroughly evaluate and develop control algorithms for robotic systems. While existing software frameworks provide some of these capabilities, integrating them into a cohesive system can be challenging and cumbersome. To overcome this challenge, we have modularized each component in \textbf{\texttt{RPC}} to ensure easy and seamless integration or replacement with new modules. Additionally, our framework currently supports a variety of model-based planning and control algorithms for robotic manipulators and legged robots, alongside essential debugging tools, making it easier for users to design and execute complex robotics tasks. The code and usage instructions of \textbf{\texttt{RPC}} are available at \href{https://github.com/shbang91/rpc}{https://github.com/shbang91/rpc}.            
    
\end{abstract}

\section{INTRODUCTION}
\label{sec:introduction}
\input{sections/introduction.tex}

\section{System Modules}
\label{sec:system_modules}
\input{sections/system_modules.tex}

\section{DEMONSTRATIONS} 
\label{sec:demonstrations}
\input{sections/demonstrations.tex}

\section{CONCLUSIONS}
\label{sec:conclusion}
\input{sections/conclusion.tex}

\addtolength{\textheight}{-12cm}   




\section*{ACKNOWLEDGMENT}
This work was supported by the Office of Naval Research (ONR), Award No. N00014-22-1-2204.

\bibliographystyle{IEEEtran}
\balance
\bibliography{references}

\end{document}

%% file: sections/introduction.tex
In order to deploy control algorithms safely and reliably on robotic hardware, it is essential to first evaluate them extensively and rigorously in simulation environments. While learning-based control algorithms are widely popular~\cite{Seo2023DeepTeleoperation, Chi2024DiffusionDiffusion}, model-based control algorithms remain essential due to their capacity to provide systematic analysis and generalization without requiring data collection~\cite{Tassa2014Control-limitedProgramming,Wensing2024Optimization-BasedRobots}. However, these algorithms are often challenging to implement due to their complex optimization processes and are typically not available in a user-friendly form for the broader technical community. Finally, debugging tools are essential within the software components for complex robots, as they are critical for diagnosing and resolving issues that arise during the development and deployment of control processes.    

To tackle these challenges, this paper introduces a software architecture designed to integrate multiple physics-based simulators, model-based planning and control modules, visualization tools, plotting and logging utilities, and operator interfaces for robotic systems. This integration facilitates intuitive deployment, thorough testing, and iterative refinement of control algorithms, significantly enhancing the reliability of the robot control deployment process. 


\subsection{Related Work}
Recent advancements in physics-based simulators, such as MuJoCo~\cite{Todorov2012MuJoCo:Control}, PyBullet~\cite{Coumans2016PyBulletLearning}, and Raisim~\cite{Hwangbo2018Per-ContactDynamics}, have significantly accelerated progress in robotics research. However, interfacing these simulators and control modules, or integrating new robots into these environments, remains a complex challenge. Each simulator provides its own APIs for low-level access to its full capabilities, requiring the development of higher-level interface (wrapper/utility) functions to enable the effective evaluation of controllers within these environments. To address this, the \textit{mc-mujoco} library~\cite{Singh2023Mc-Mujoco:MuJoCo} facilitates integration between MuJoCo and the \textit{mc-rtc} robot control framework~\cite{Mc-rtc}, while the \textit{PnC} library~\cite{Ahn2021VersatileRobots} bridges DART~\cite{Lee2018DART:Toolkit} with its associated control framework. Compared to these libraries, our software framework offers greater versatility by supporting interfaces with both the MuJoCo and PyBullet simulators.         

Significant research has been devoted to model-based motion planning and control algorithms for robotic systems, yet much of the related software remains proprietary, difficult to access, or challenging to integrate and test. While some libraries, such as the inverted pendulum-based gait planner~\cite{Caron2020Capturability-BasedHeight} and the task space inverse dynamics controller~\cite{DelPrete2016ImplementingSensors}, have been released as open-source, they are not self-contained and require external libraries to complete the motion planning and control pipeline.     
In contrast, libraries like \textit{OCS2}~\cite{FarbodFarshidianandOthersOCS2:Systems}, \textit{Drake}~\cite{drake}, \textit{open-robotics-software}~\cite{IHMCRobotics2018IHMCSoftware}, \textit{Cheetah-software}~\cite{Kim2019HighlyControl}, and \textit{PnC}~\cite{Ahn2021VersatileRobots} offer integrated solutions that include both motion planning and feedback control, along with a comprehensive test environment. However, these solutions have significant drawbacks: \textit{OCS2} is heavily reliant on ROS, \textit{Drake} lacks versatile options regarding simulators and visualization tools, \textit{open-robotics-software} is Java-based, and both the \textit{Cheetah-software} and \textit{PnC} provide limited planning and control options. Our proposed framework overcomes these limitations by being ROS-independent, C++-based for real-time performance, and highly versatile.    

Visualizing robots alongside their controller states and planned actions is essential during code development and debugging. Additionally, a user-friendly operator interface can minimize delays in non-autonomous operations and expand the range of possible tasks. To meet these needs,~\cite{Johnson2015TeamTrials} developed debugging tools that include logging features, a robot visualizer, and a practical user interface (UI). Similarly,~\cite{Howell2022PredictiveMuJoCo} introduced a graphical user interface (GUI) that provides an interactive simulation environment with live plotting and user-configurable parameters. Our framework offers comparable capabilities, featuring comprehensive visualizations, live plotting, logging, and operational tools that support both development and deployment---capabilities extending our previous work~\cite{Ahn2021VersatileRobots}.

\subsection{Contributions} The main contributions of this paper are the following:
\begin{enumerate}
    \item We have devised a lightweight yet comprehensive open-source software framework, named \textbf{\texttt{RPC}}, which integrates motion planning and control modules for robotic applications, along with graphical tools for robot visualizations, data logging, and user interfaces.
    \item We provide a versatile software interface that allows the proposed framework to be easily deployed across multiple high-fidelity physics simulators for extensive algorithm testing and demonstrate its smooth integration within the ROS environment for conducting hardware experiments.
    \item We demonstrate the use of the proposed framework in loco-manipulation tasks performed by the humanoid robot DRACO 3 in both simulated and real hardware environments.
\end{enumerate}

\subsection{Organization}
The remainder of this paper is organized as
follows. Section~\ref{sec:system_modules} provides a concise overview of the proposed software framework and its key modules. Section~\ref{sec:demonstrations} demonstrates the practicality of the framework through humanoid locomanipulation tasks. Finally, Section~\ref{sec:conclusion} concludes the paper and discusses potential directions for future work.

%% file: sections/system_modules.tex
\begin{figure*}[t!]
    \centering
    \includegraphics[width=0.93\textwidth]{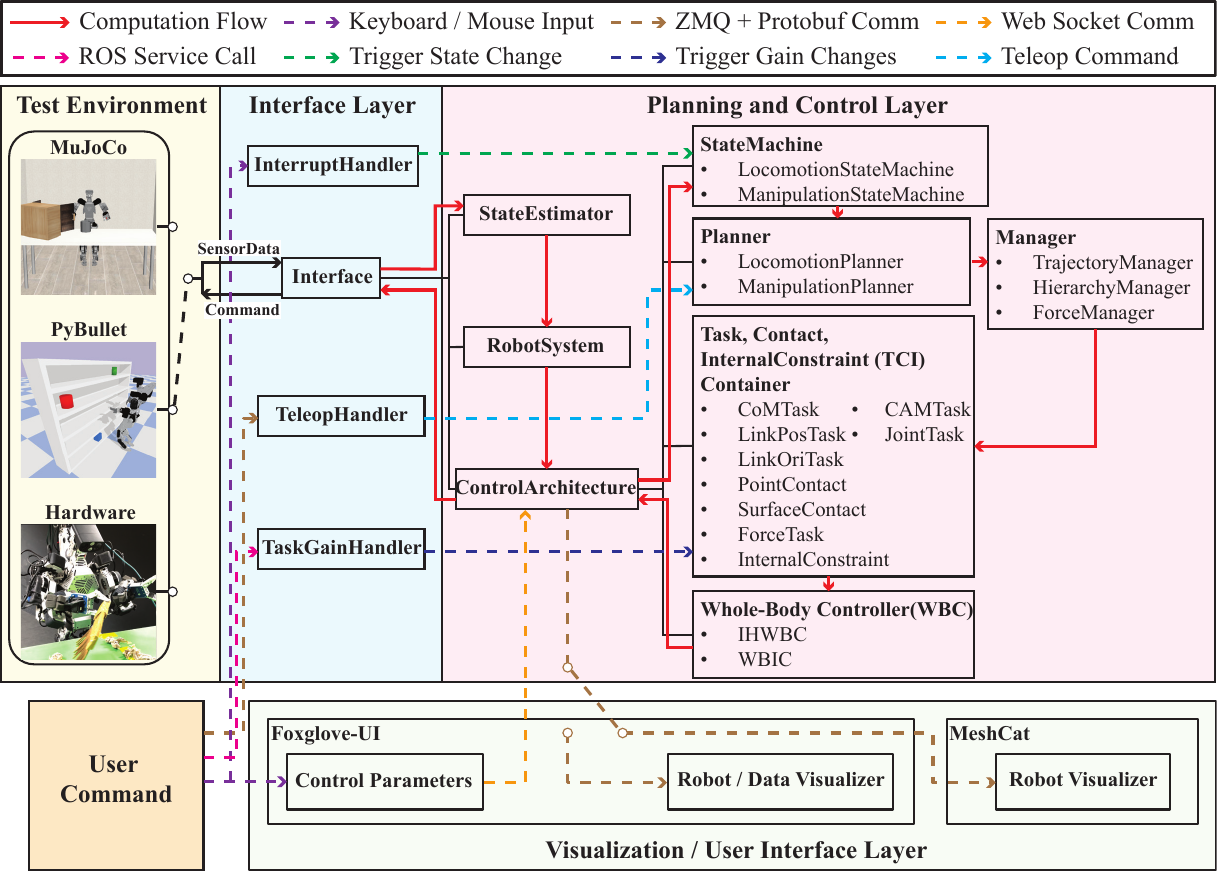}
    \caption{\textbf{Overall software architecture:} \textbf{\texttt{RPC}} consists of the Test Environment, Interface Layer, Planning and Control Layer, and Visualization / User Interface Layer. Each layer includes several modules, and their interaction methods are illustrated in different line types.} 
    \label{fig:software_architecture}
\end{figure*}

This section describes the overall software framework of \textbf{\texttt{RPC}} and its key modules.  An overview of the proposed software architecture is illustrated in 
Fig.~\ref{fig:software_architecture}.

\subsection{Test Environment}
In \textbf{\texttt{RPC}}, we integrate two high-fidelity physics simulators: PyBullet and MuJoCo. This allows us to readily evaluate control algorithms across multiple simulators (e.g., to asses the controller's performance under different contact dynamics models) to enhance the versatility and reliability of the resulting algorithms. Each robot has its associated main simulation script, where our utility functions --- integrated with the APIs of PyBullet and MuJoCo --- allow for reading sensor data from the simulation and applying control signals to the robot's actuators. 


\subsection{Interface Layer}
The modules in this layer enable communication between the low-level controllers in the \textbf{Test Environment} and the high-level layers (i.e., \textbf{User Command} and the \textbf{Planning and Control Layer}). The versatility of these modules ensures that our framework can be seamlessly applied in both physics simulators and hardware environments.

The key modules of this layer are the following:
\begin{itemize}    
    \item \textbf{InterruptHandler}: This module manages state transitions within either the \texttt{LocomotionStateMachine} or \texttt{ManipulationStateMachine} based on user input commands. In our current setup, these commands are sent via keyboard, but other input devices (e.g., joysticks) can be easily added. 

    \item \textbf{Interface}: This module manages the communication of sensor data and commands from the \textbf{Test Environment} to the \textbf{Planning and Control Layer}. In one direction, the \texttt{SensorData} read from the \textbf{Test Environment} is updated at each servo loop and relayed to the \texttt{StateEstimator} class to update the robot's states. This data includes measurements from the IMU, joint encoders, F/T sensors, and cameras. In the opposite direction, the \texttt{GetCommand} function is invoked to compute the \texttt{Command} using the \textbf{Planning and Control Layer}. This \texttt{Command}, which consists of joint positions, velocities, and torques, is then applied to the robot's actuator via a joint impedance controller.      
    \item \textbf{TeleopHandler}: This module manages the communication with the teleoperation devices (e.g., RealSense T265 camera) through ZeroMQ~\cite{Hintjens2013ZeroMQ:Applications} and Protocol Buffers~\cite{Google2008ProtocolBuffers}, and relays the teleoperation commands to the \texttt{ManipulationPlanner}. 

    \item \textbf{TaskGainHandler}: This module manages the task PD controller gains in the \texttt{TCIContainer} during both simulation and hardware operation. Users can adjust the task gains in real-time through a user-specified communication protocol. This is particularly useful for hardware experiments that require frequent gain tuning. In the current implementation this is achieved using ROS messages and ROS service calls.  

\end{itemize}

\subsection{Planning and Control Layer}
The \textbf{Planning and Control Layer} is a crucial component that enables robots to perform complex tasks reliably and efficiently. The modules contained in this layer work together to transform high-level goals into precise, coordinated low-level actuator commands. The key components include \texttt{StateEstimator}, \texttt{RobotSystem}, \texttt{ControlArchitecture}, \texttt{StateMachine}, \texttt{Planner}, \texttt{Manager}, \texttt{TCIContainer}, and \texttt{WBC}. Their flexibility allows for easy adaptation to new robots, increasing the versatility of our software framework. 

\begin{itemize}
    \item \textbf{StateEstimator:} This class estimates the floating base state (i.e., SE(3) and twist). We provide two different state estimators: 1) a simple estimator that uses only an IMU and leg joint encoders~\cite{Bang2023ControlBody}, and 2) a KF-based estimator~\cite{Flayols2017ExperimentalRobots}.     
    
    \item \textbf{RobotSystem:} This class serves as a wrapper around Pinocchio~\cite{Carpentier2019TheDerivatives}, enabling efficient computation of rigid body dynamics. It provides APIs for obtaining kinematic and dynamic properties of the robot, such as link Jacobians, centroidal states~\cite{Orin2013CentroidalRobot}, and the mass matrix. This class is instantiated using a universal robot description file (URDF) and is updated in each control loop to reflect the robot's current configuration and velocity states in conjunction with the chosen \texttt{StateEstimator}.      
    
    \item \textbf{ControlArchitecture:} This class is a key component of \textbf{\texttt{RPC}}, responsible for generating control commands that are passed to \texttt{Command} by integrating all necessary modules for robot planning and control. It is instantiated with a \texttt{RobotSystem} object and, within 
 its constructor, instances of the \texttt{StateMachine}, \texttt{Planner}, \texttt{Manager}, \texttt{TCIContainer}, and \texttt{WBC} classes are created. During each control loop, the \texttt{GetCommand} function is invoked to compute the control commands (i.e., joint positions, velocities, and torques). 
    
    \item \textbf{StateMachine:} This class coordinates the robot's complex behaviors by managing a finite number of locomotion or manipulation states and by handling transitions between them based on predefined conditions or inputs. Each state corresponds to a distinct contact mode or a specific task contributing to a modular and structured control system. 
    
    This \texttt{StateMachine} class is instantiated with the \texttt{ControlArchitecture} and \texttt{RobotSystem} instances. At the start of each state, the \texttt{FirstVisit} function is invoked to initialize the desired control signal trajectories, typically using predefined temporal parameters, via the \texttt{Planner} or \texttt{Manager} in the \texttt{ControlArchitecture}. During each control loop, the \texttt{OneStep} function updates these trajectories as defined in \texttt{FirstVisit}. The \texttt{EndOfState} function checks whether the termination conditions for the current state have been met and triggers a state transition based on predefined temporal parameters, contact events, or signals from the interrupt handler.  

    \item \textbf{Planner:} The \texttt{Planner} module is responsible for generating motion plans that enable robots to move effectively and efficiently. Within \textbf{\texttt{RPC}}, we have implemented both locomotion and manipulation planners to support robotic systems. 
    
    To address the different requirements for balance and stability during walking, we provide two types of locomotion planners: 1) the Divergent Component of Motion (DCM) planner~\cite{Englsberger2015Three-DimensionalMotion} and 2) the convex Model Predictive Control (MPC) planner~\cite{DiCarlo2018DynamicControl}. The DCM planner is designed for quasi-static walking, utilizing DCM dynamics based on the Linear Inverted Pendulum Model (LIPM) to calculate the robot's motions. It takes a pre-determined sequence of foot placements (generated by a footstep planner) as input and generates a DCM trajectory, which serves as the reference signal for the robot's center of mass (CoM) task in \texttt{TCIContainer}. 
    
   On the other hand, the convex MPC planner is optimized for dynamic walking, employing a Single Rigid Body Dynamics (SRBD) model. The MPC planner takes velocity commands---specifically, the desired CoM velocity in the x and y directions and the yaw velocity of the torso---as input. It then generates a ground reaction force (GRF) trajectory, which is used as the reference for the foot force task in \texttt{TCIContainer}. Additionally, we provide a variant of the MPC planner known as VI-MPC~\cite{Bang2024VariableManeuvers}, which extends the SRBD model by incorporating composite rigid body inertia. This enhancement allows the planner to compute more reliable GRFs, enabling faster and more efficient maneuvers.           

   For manipulation planning, we provide various interpolation methods for trajectory generation to ensure smooth motions as the robot's arm reaches a target position. The following interpolation options are included in \textbf{\texttt{RPC}}: \texttt{CosineInterpolate}, \texttt{HermiteCurve}, \texttt{MinJerkCurve}, and \texttt{CubicBezier}.  

    \item \textbf{Manager:} The \texttt{Manager} module is a utility class that serves as an interface between the \texttt{Planner} and \texttt{TCIContainer}. Specifically, it receives the desired control signals from the \texttt{Planner} and updates the corresponding \texttt{Task} or \texttt{Contact} elements in the \texttt{TCIContainer}.  
    
    \item \textbf{TCIContainer:} The \texttt{TCIContainer} module is designed to  
    efficiently manage a list of \texttt{Task}, \texttt{Contact}, and \texttt{InternalConstraint} elements, which are frequently updated for use in the \texttt{WBC} module. This module initializes and maintains these lists, ensuring that they are easily accessible and modifiable. We also provide modular \texttt{Task}, \texttt{Contact}, and \texttt{InternalConstraint} classes, making them reusable across different types of \texttt{WBC} instances and flexible enough to accommodate robot-specific tasks (e.g., Whole-body Orientation task~\cite{Chen2023IntegrableRobots}). The following options are available: \texttt{JointTask}, \texttt{SelectedJointTask}, \texttt{LinkPosTask}, \texttt{LinkOriTask}, \texttt{CoMTask}, \texttt{CAMTask}, \texttt{WBOTask}, \texttt{PointContact}, \texttt{SurfaceContact}, \texttt{RollingJointConstraint}.    

    \item \textbf{WBC:} The \texttt{WBC} module is designed to convert the high-level task specifications (such as CoM or End-effector task objectives) into low-level motor commands that drive the robots. To address the different control requirements, we provide two types of \texttt{WBC}: 1) the Implicit Hierarchical Whole-body Controller (IHWBC)~\cite{Ahn2021VersatileRobots} and 2) the Whole-body Impulse Controller (WBIC)~\cite{Kim2019HighlyControl}.

    IHWBC employs an implicit hierarchy between the \texttt{Task} and soft constraints to handle \texttt{Contact}, enabling smooth task and contact transitions. This controller is suitable for robots with highly accurate dynamics models as it calculates joint position and velocity commands in a dynamically consistent manner via integration schemes~\cite{IHMCRobotics2018IHMCSoftware}.

    In contrast, WBIC employs a strict hierarchy between the \texttt{Task} and hard constraints to handle \texttt{Contact} using a null-space projection technique to strictly hold task priority. This WBC is effective for robots that need reliable motion stabilization through joint position feedback control since it utilizes an inverse kinematics algorithm to compute joint position, velocity, and acceleration commands. 
    
\end{itemize}
    \begin{figure*}[t!]
    \centering
    \includegraphics[width=\textwidth]{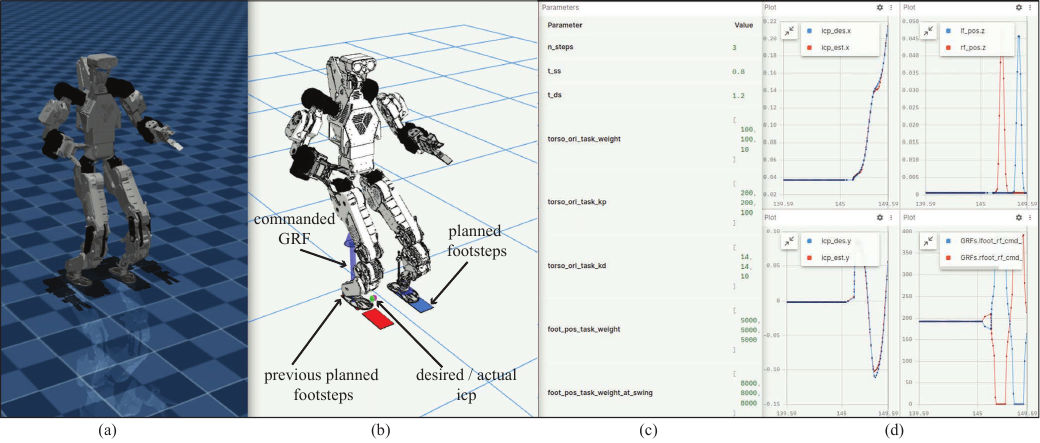}
    \caption{\textbf{Foxglove UI usage:} (a) MuJoCo simulation. (b) Robot model visualization window in Foxglove. (c) Control parameters tuning window in Foxglove. (d) Data visualization window in Foxglove.} 
    \label{fig:foxglove_ui}
\end{figure*}

\subsection{Visualization / User Interface Layer}
We have set up two main forms of operating and visualizing robots: 
one where the user sends commands via the keyboard and can visualize the robot via Meshcat, 
and another one using a GUI to operate and visualize the robot via Foxglove.
\begin{itemize}
    \item \textbf{Meshcat:} Meshcat is a 3D viewer that communicates over websockets and runs in the browser. 
    We use the Python bindings for Meshcat~\cite{Meshcat-python:Viewer} to host the \texttt{Robot Visualizer} server and 
    visualize several elements, such as the overall robot, 
    its DCM, its CoM, its desired GRF's, and its planned footsteps. 
    This data is sent out as ZMQ messages and is also stored in a pickle file. 
   This last step allows us to replay any given log in greater detail, including all visual elements, while scrolling through time.
    
    \item \textbf{Foxglove:} Foxglove~\cite{FoxgloveDevelopers2024Foxglove} is a software package equipped with visualization, plotting, logging, and operation capabilities for robotic platforms. 
    Currently, it easily integrates into ROS systems, but requires additional steps to integrate 
    into other environments. Hence, we have developed several modules to use
    it without requiring ROS in environments such as \textbf{\texttt{RPC}}. 
    In particular, we u two different servers by utilizing the Foxglove websocket protocol: the \texttt{Control Parameters} server and the \texttt{Robot/Data Visualizer}. 
    The \texttt{Control Parameters} server, shown in Fig.~\ref{fig:foxglove_ui}(c), provides the user with a web GUI to adjust the robot’s control parameters while operating the robot. A client that subscribes to these parameters is run in the \texttt{ControlArchitecture} and updates its values accordingly.
    The \texttt{Robot/Data Visualizer} hosts topics rendered in the 3D viewer, shown in Fig.~\ref{fig:foxglove_ui}(b), and/or in plots, shown in Fig.~\ref{fig:foxglove_ui}(d). We also provide the tools to log this data into an MCAP file, which can then be uploaded to Foxglove for synchronized replay, complete with all its visual elements.
    The ability to run either server independently of one another allows the user to implement alternative visualizers without losing the ability to adjust robot parameters while in operation.
\end{itemize}

%% file: sections/demonstrations.tex
In this section, we demonstrate the practicality of our proposed framework for locomanipulation tasks for a 25-DOF humanoid robot, DRACO 3~\cite{Bang2023ControlBody}, in both simulation and hardware environments. For more details on the experiments conducted in this section, please refer to the accompanying video. 

\begin{figure*}[t]
    \centering
    \includegraphics[width=\textwidth]{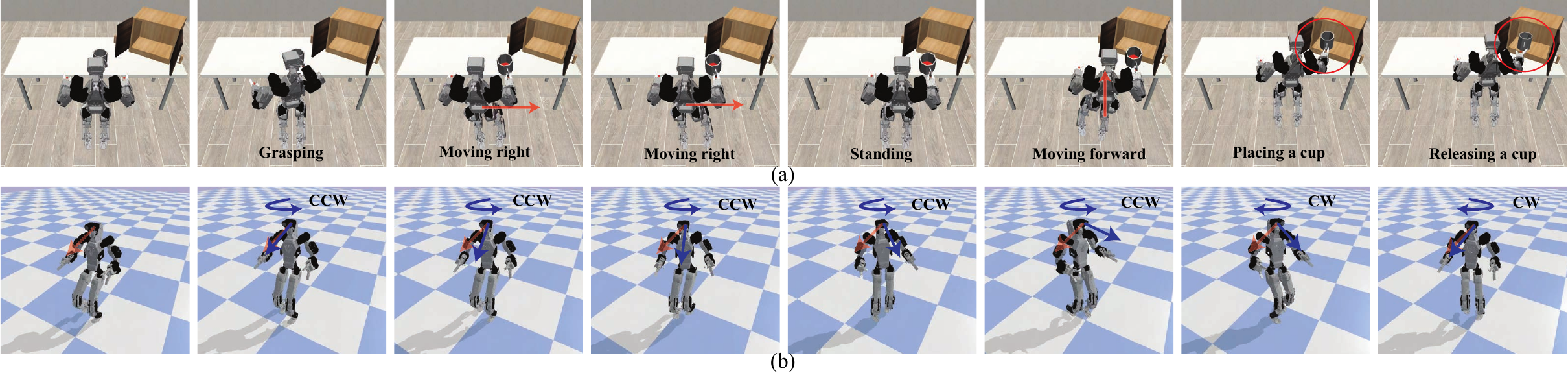}
    \caption{\textbf{Simulation snapshots:} (a) Teleoperation-based locomanipulation for a cup shelving task. (b) Convex MPC-based omnidirectional walking task. Red arrows represent the initial heading, and the blue arrows indicate the current heading.} 
    \label{fig:simulation_demonstration}
\end{figure*}
\subsection{Simulation}
We demonstrate here our ability to synthesize complex manipulation and locomotion behaviors in simulation environments using \textbf{\texttt{RPC}}.

First, we performed a locomanipulation task in MuJoCo, where the robot was required to complete a cup shelving task, as shown in Fig.~\ref{fig:simulation_demonstration}(a). In this task, the manipulation motion plan (robot's right hand SE(3) pose) was provided through teleoperation by an operator using the Realsense T265 tracking camera, with gripper commands (open and close) triggered by keyboard keystrokes. These manipulation commands were relayed to the \texttt{ManipulationPlanner} via the \texttt{TeleopHandler}. For the locomotion planner, we employed the DCM planner with predefined temporal parameters and step lengths. The footstep plan was triggered by keyboard keystrokes through \texttt{InterruptHandler}. Given these manipulation and locomotion plans, IHWBC computed joint commands in each corresponding \texttt{StateMachine}, which were then applied to the actuators via \texttt{Command} to accomplish the task. Notably, the robot was able to freely move the cup while holding it, regardless of its locomotion state, due to our architecture's use of independent \texttt{StateMachine} instances for manipulation and locomotion tasks, unlike the approach in~\cite{Singh2023Mc-Mujoco:MuJoCo}.

Next, we performed a dynamic locomotion task in Pybullet, where the robot was required to maneuver omnidirectionally, as shown in Fig.~\ref{fig:simulation_demonstration}(b). In this task, given velocity commands from an operator through \texttt{InterruptHandler}, the SRBD model-based convex MPC planner generated the reference GRF trajectory, while the Raibert hueristics~\cite{Raibert1983StableLocomotion} was used to decide the desired foot placement. Based on these locomotion plans, WBIC computed joint commands to track the body posture, swing foot pose, and GRF.  

\begin{figure*}[t]
    \centering
    \includegraphics[width=\textwidth]{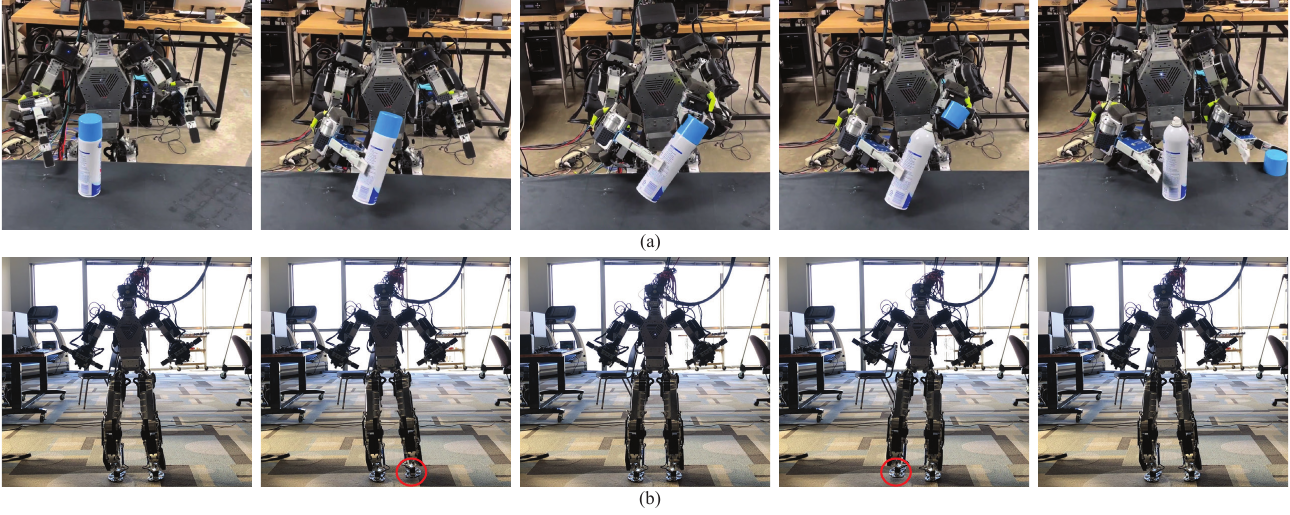}
    \caption{\textbf{Hardware experiment snapshots:} (a) Teleoperation-based bi-manipulation for a spray cap removal task. (b) DCM-based in-place quasi-static stepping.} 
    \label{fig:hardware_demonstration}
\end{figure*}

\subsection{Hardware}
DRACO 3 utilizes ROS as its middleware, so we set up a ROS nodelet where \textbf{\texttt{RPC}} receives joint state information and transmits joint commands via shared memory. Within the nodelet modules, \texttt{Interface} effortlessly connects the \textbf{Planning and Control Layer} to the robot without any modifications, ensuring a smooth and reliable transition between simulation and real hardware. 

First, we performed a spray cap removal task as shown in Fig.~\ref{fig:hardware_demonstration}(a), where the robot was required to grasp a spray can with one hand and remove its cap with the other hand while balancing on its feet. Since this task required bimanual manipulation, we used a VR device (Meta's Oculus Quest 2) for the robot's teleoperation. The underlying software architecture for this task is similar to the cup shelving task described in the previous subsection.   

Then, we also performed a DCM-based stepping-in-place task, as shown in Fig.~\ref{fig:hardware_demonstration}(b). In this task, we employed the IHWBC controller to track body posture, DCM, and swing foot pose. 

It is important to note that these hardware experiments were extensively tested in simulation environments before being evaluated on the real hardware.

%% file: sections/conclusion.tex
In this work, we develop and open-source \textbf{\texttt{RPC}}, a testing and development control software framework designed for complex robotic systems, with a particular focus on humanoid robots. The framework is fully equipped with simulators, planning and control modules, and debugging tools, enabling robotics researchers to develop and test their algorithms with minimal external dependencies. Its modular design ensures that the software architecture can be easily extended or adapted to incorporate new simulators, hardware platforms, or control strategies, thereby enhancing the framework's flexibility and scalability. We believe that \textbf{\texttt{RPC}} will greatly facilitate the development, testing, and deployment of advanced robotics systems.